\documentclass[sigconf]{acmart}
\AtBeginDocument{%
  \providecommand\BibTeX{{%
    \normalfont B\kern-0.5em{\scshape i\kern-0.25em b}\kern-0.8em\TeX}}}

\setcopyright{acmcopyright}
\copyrightyear{2019}
\acmYear{2019}
\acmConference{DMAIC'19}{August 5}{Anchorage, AL}
\acmBooktitle{DMAIC'19: KDD19 Workshop on Data Mining and AI for Conservation, August 5, 2019, Anchorage, AL}
\acmISBN{ }

\begin{document}
\balance
\title{Animal Wildlife Population Estimation Using Social Media Images Collections}


%\email{webmaster@marysville-ohio.com}
%\affiliation{%
%  \institution{Institute for Clarity in Documentation}
%  \streetaddress{P.O. Box 1212}

\author{Matteo Foglio}
\affiliation{%
  \institution{Politecnico Milano}
  \streetaddress{}
  \city{}
  \state{}
  \postcode{}
}
\email{foglio.matteo@gmail.com}
%\email{mfogli2@uic.edu}

\author{Lorenzo Semeria}
\affiliation{%
  \institution{Politecnico Milano}
  \streetaddress{}
  \city{}
  \state{}
  \postcode{}
}
\email{lor.semeria@gmail.com}

\author{Guido Muscioni}
\affiliation{%
  \institution{Politecnico Milano}
  \streetaddress{}
  \city{}
  \state{}
  \postcode{}
}
%\email{gmusci2@uic.edu}
\email{guido.muscioni@mail.polimi.it}

\author{Riccardo Pressiani}
\affiliation{%
  \institution{Politecnico Milano}
  \streetaddress{}
  \city{}
  \state{}
  \postcode{}
}
%\email{rpress4@uic.edu}
\email{riccardo.pressiani@mail.polimi.it}

\author{Tanya Berger-Wolf}
\affiliation{%
  \institution{University of Illinois at Chicago}
  \streetaddress{}
  \city{}
  \state{}
  \postcode{}
}
\email{tanyabw@uic.edu}


% This to avoid "All authors" being included in the authors
\renewcommand{\shortauthors}{Foglio, et al.}

\begin{abstract}
We are losing biodiversity at an unprecedented scale and in many cases, we do not even know the basic data for the species. Traditional methods for wildlife monitoring are inadequate. Development of new computer vision tools enables the use of images as the source of information about wildlife. Social media is the rich source of wildlife images, which come with a huge bias, thus thwarting traditional population size estimate approaches. Here, we present a new framework to take into account the social media bias when using this data source to provide wildlife population size estimates. We show that, surprisingly, this is a learnable and potentially solvable problem.

\end{abstract}




\maketitle

\section{Introduction}
% the need: 60% wildlife, wildlife population monitoring, issues with traditional methods
% warning: biases, naive social media usage
% the opportunity: wildbook
According to the recent UN report~\cite{UNreport}, we are losing biodiversity at an unprecedented scale and rate. In many cases, we do not have the basic data for the species, such as population sizes. Traditional methods for  wildlife monitoring are expensive, unsustainable, and unscalable~\cite{witmer2005wildlife}.  Alternative approaches are urgently needed.

Development of new computer vision tools for wildlife, such as Wildbook\texttrademark~\cite{berger2017wildbook}, allows to analyze a huge quantity of animal images and to identify animals at the individual level. This identification process is fundamental for the use of several wildlife estimation techniques, such as capture-mark-recapture~\cite{lancia2005estimating}. This enables the use of images as the source of information about wildlife. 
%TOSHORTEN Social media is the rich source of information about many aspects of our life, including the wildlife that we collectively observe. However, the use of social media data is not exempt from challenges \cite{olteanu2016social}. 
People upload images according to their tastes and experiences, creating biases which have been proven to affect wildlife population estimates. 
% TO-SHORTEN Traditional wildlife estimation methods have been developed under the assumption that the data used must satisfy precise hypothesis \cite{lancia2005estimating}, which are rarely satisfied when using social media as a data source. 
Therefore, it is necessary to develop new frameworks to take into account the social media bias when using this data source in providing wildlife estimates. \cite{olteanu2016social} %Here, we propose just such a framework.

%\section{Related Work}
% single images instead of images collection
Recently, work has started to leverage social media data for animal wildlife population size estimation \cite{menon2016animal,menon2017animal}. This work has focused on predicting the likelihood of an image being shared on social media. 
% TOSHORTEN This work has focused on predicting the likelihood of an image being shared on social media, which was then used to compute the probability for an animal to appear on social media and then to understand how the animals' appearing in online pictures were related to real animal population. %Part of our research is focused on the inverse process.
Our goal is to directly estimate the number of animals photographed by a social media user, given the pictures they shared online. These two numbers may differ if the user posts online only a subset of pictures taken.
Moreover, we elevate the problem from single images to image collections since we believe that the shareability of an image is a function not only of the image features, but also of the features of the other photos taken at the same event.

\section{Problem Statement}
Let $\mathcal{C}_i$ be the set of images from a single image collection posted on a social media platform. Let $\mathcal{SD}_i$ be the set of images in one or more SD cards from which the images in $\mathcal{C}_i$ have been taken.

First, given an images collection $\mathcal{C}_i$ shared on social media by a photographer $P_i$, we want to estimate $\hat{N}_i$ for the real number of animals $N_i$ present in the set $\mathcal{SD}_i$ of the SD cards of the photographer $P_i$. We refer to this problem as the {\sc estimate problem}. We model it as a regression model: given an images collection $\mathcal{C}_i$, we map it to a set of features $\mathcal{F}_i$ which includes the number of individual animals $n_i$ in the collection $\mathcal{C}_i$. Our regression problem is then to predict $N_i$ from $\mathcal{F}_i$.
Then, using a set of estimates $\mathcal{N}_i$, we want to provide an estimate for the number of animals of the given species. 

Secondly, given the collection of pictures $\mathcal{SD}_i$ that were taken by the photographer, in addition to knowing which were shared, we define our problem as a binary classification. The labels are ``shared" and ``not shared". %%TOSHORTEN, for images from $\mathcal{C}_i$ or $\mathcal{T}_i \cap \neg \mathcal{C}_i$ respectively. 
We refer to this part of the problem as
{\sc shareability problem}.

\section{Datasets}
%\subsection{GGR1 and GGR2}
% more explicitly states that we have a ground truth?
\noindent{\bf GGR1 and GGR2.} For our research, we focus on Grevy's zebras (\textit{Equus grevyi}) for two reasons. First, its population lives in a delimited area \cite{rubenstein2016equus} which makes it possible to census it. % spreads from the central part of Kenya to Southern Ethiopia .
Most recent such census were done using photographs from volunteer citizen scientists over a course of 2 days in 2016 and in 2018: the Great Grevy's Rally~\cite{berger2016great}. %Volunteers cover the entire habitat of the species taking pictures of Grevy's zebra.
Wildbook\texttrademark was then used to analyze the images and identify individual animals, resulting in two datasets GGR1(2016) and GGR2(2018), which have also been used to provide the ground truth for the population size.
%, each photo has been analyzed and animals of different species have been detected. Moreover, the software provides biological information for individuals belonging to some supported species: Grevy's zebra, Plains zebra, Reticulated giraffe and Masai giraffe. The results are two datasets of thousands of images taken during the GGR1 in 2016 and the GGR2 in 2018. These datasets have also been used to provide the ground truth for the population size.

%\subsection{Flickr\texttrademark}
\noindent{\bf Flickr\texttrademark} was used as the social media data source. % to provide Grevy's zebra estimates for different years. 
Using API, we downloaded images matching ``grevy's zebra" keyword, together with their albums. %Using a Python implementation of Flickr's API \cite{flickrAPIpython}, we searched for pictures using the function flickrObj.photos.search by passing "grevy's zebra" as text parameter. Then, we filter out all those images which were not taken in the wild. Finally, for each image we downloaded its album.

\section{Methods}
% mturk, features, time series, regression, jolly-seber
%TOSHORTEN The problem of estimating the number of animals of a given species using social media images presents two main challenges. First, the estimation of the number of animals photographed by a user but not posted on social media. Second, the integration of the estimate with traditional biological techniques.
%First, the estimation of the bias related to the shareability of images collection. 

%\subsection{Dataset labelling through crowdsourcing marketplace}

For the {\sc estimate problem}, we lacked the ground truth of people's preferences in sharing images. Using the images from GGR1 and GGR2 we created an online survey for each of the SD card collected during the event. The survey contained a list of all the photos of the SD card, and for each of these photos we asked the interviewee whether they would share it on social media or not.

From each image we extracted a set of features $\mathcal{F}_i$ to describe various aspects of its beauty as well as biological information regarding the animals in the photo, similar to~\cite{menon2017animal}.  We used Wildbook\texttrademark  to identify each individual animal in all the photos. We have also introduced features modeling the structure of the source collection, in order to account for all the pictures that were taken. % For each individual, Wildbook provides also the sex, the age as well as the side that the animal shows in the photo.
%
% TOSHORTEN For the estimate problem, single image features have then been aggregated into time series to model an image collection in its entirety. %Each time series has been obtained by listing the values of a certain single-image feature sorted by the order in which the images appear in the collection. The sampling rate has been considered constant since we wanted to model the relative order of the picture in the SD card rather than their distribution over absolute time. This is because we believe that an user decides which images to share also by comparing each image to its adjacent in the SD card.
%
%TOSHORTEN For the binary classification problem we have also introduced features modeling  the structure of the source collection, such as the file name number, in order to account for all the pictures that were taken. % These new features aim at more closely tracking the choices that a user makes when selecting what images they will share on social media.
%Moreover, we developed controlled experiments to test the statistical significance of some of the features.% on the shareability of images. In particular,
% TOSHORTEN Moreover, by developing controlled experiments, we studied the effects of the number of animals and individual animals in an images collection, the side shown by animals, and the size of the animals in the picture and possible differences in the shareability of among different individuals.

Using our labelled pictures from GGR1 and GGR2, we trained a regression model to predict the percentage of animals that the interviewee would  share. The inverse of this number is a coefficient $k_i$ so that the estimated number of animal photographed by the user $P_i$ is $\hat{N}_i = k_i * n_i$. This model has then been applied to each images collection $C_\mathcal{}{i}$ retrieved from Flickr\texttrademark. 

In order to provide an estimate for the entire population, we relied on the Jolly-Seber method~\cite{jolly1965explicit,seber1965note}, a capture-mark-recapture technique that uses multiple years' data. %Even if through Wildbook\texttrademark\, we knew the number of animals recaptured among different years on social media, we lacked this data for the animals in the SD card of each user. In fact, for each images collection $\mathcal{C}_i$, the regression model provides an estimate $\hat{N}_i$ for the number of animals using the coefficient $k_i$, but not about the identity of each of this animals. 
%In other words, we need a multiplier $k_{rec}$ to estimate the number of recaptured animals $R_{i,j}$ among the different SD cards so that we can have an estimate $\hat{R}_{i,j} = k_{rec_{i,j}} * r_{i,j}$ where $r_{i,j}$ is the number of recaptured animals on social media among the two images collection $C_\mathcal{i}$ and $\mathcal{C}_j$. 
Given two years $year_m$ and $year_n$, we took all the coefficients $k_i$ coming from images collections of photos taken in those two years, and then estimated the overall coefficient using the average of these coefficients: $k_{rec} = Avg_{\{i \in \{year_m \cup year_n\}\}}(k_i)$. Having all the variables for the Jolly-Seber method we provided an estimate for the Grevy's zebra population. 
Finally, we multiply this estimate by the computed coefficient to take into account the recaptures which cannot be detected by a computer vision tool due to the lack of symmetry in Grevy's zebra stripes on either side. %In fact, if the same zebra appears in two pictures, the probably $P$ of identifying it as the same individual is the probability that the animal shows the same side in both pictures.
%\begin{equation}
%P = \sum_{i \in N }^{}{P(side_{i-t_{k_1}}) *  P(side_{i-t_{k_2}})  }
%\end{equation}

%\section{Understanding behavior: experiments}
%\textcolor{red}{Sezione aggiunta da Lorenzo}
%Our classifier identified some features to be of greater importance when learning the shareability of an image, as we discussed previously. 
%The features the were considered important can be further tested to show if they have an actual effect on the sharing rate of images. 
%For this reason we design repeated experiments to measure the effect of individual features.
%We chose to focus on two features: number of animals in the collection and position of the image in the collection

%We created surveys with a limited number of images with similar values for all features but the one being tested, and surveyed users on Mechanical Turk to measure the effect of the feature in the sharing behavior.

%For the position of the image we created surveys containing the same images, chosen to be similar within the collection, and shuffled them to generate multiple surveys with the same pictures in different position. 

%Testing the number of animals can be achieved by altering the number of images, the average number of animals per picture, or both. 
%For this reason we created surveys having 
\section{Experimental Setup}
For the estimate problem we chose as weak classifier the mean and the mode, and as strong classifiers elastic net, gradient boosting trees and SVR. The models have been evaluated using a 10 times 10-fold cross-validation and R2 and MSE metrics.

For the shareability problem, we tested the performance of the state-of-the art models: Logistic Regression, K-Nearest Neighbors, Decision Trees, Random Tree Forests, AdaBoost and XGB). %TOSHORTEN Where applicable, we also performed feature selection, scaling and normalization.
The models were evaluated using 10-fold cross validation and we aimed at maximizing the F-1 score and the Accuracy.
GGR1 and GGR2 have different structures and we anticipated that the former, being more regular, would be easier to learn for our models. 
For this reason we trained on our models using either dataset, the combination of the two, and GGR1 as training set, tested on GGR2. 
%We designed these different tests to understand how the different complexity of the dataset can affect the model performance.

\section{Results}
eXtreme Gradient Boosting (XGB)~\cite{chen2016xgboost} provided the best results in both problems. %, outperforming other available models in our tests. 
For the estimate problem, it achieved a performance of R2 of 0.417, with a standard deviation of 0.155, and an MSE of 0.070, with a standard deviation of 0.020, whereas the baselines scored a negative R2 value. Results are shown in Table \ref{tab:estimates}.
For the shareability problem, this classifier achieved an Accuracy of $0.82$ with a F-1 Score of $0.62$ on the combined dataset, and an Accuracy of $0.92$ (F-1 score of $0.77$) on GGR1 alone.

\begin{table}[htb]
\vspace{-3mm}
  \caption{Estimates for Grevy's zebra population}
  \label{tab:estimates}
\resizebox{.9\columnwidth}{!}{%
  \begin{tabular}{ccccl}
    \toprule
    Year&Official&Our Approach&Jolly-Seber&Images\\
    \midrule
    2011 & 2827  & 2341 & 2069 & 3654 \\
    2012 & 1897+ & 1576 & 1063 & 3198 \\
    2013 & N/A   & 592  & 3064 & 4418 \\
    2014 & N/A   & 2522 & 600  & 2928 \\
    2015 & N/A   & 528  & 1433 & 3449 \\
    2016 & 2250  & 613  & 60   & 2323 \\
    2017 & 1627+ & 67   & 0    & 1325 \\
    \hline
    RMSE & - & 1194 & 1612 & - \\
  \bottomrule
\end{tabular}}
\vspace{-5mm}
\end{table}

\section{Conclusions and Future Work}
We proposed a species-independent framework to estimate the size of a wildlife population using social media images.
Moreover, we demonstrated that image shareability problem is learnable and that the structure of the original collection provides useful information when modeling user behavior. %Furthermore, we have shown that \textemdash unsurprisingly \textemdash a more complex dataset is harder to learn. 
%TO-SHORTEN Interestingly, we showed that using a combination of structured and less structured datasets can boost our model's performance. The fact that the information retrieved from the structured dataset is usable also for a more complex dataset confirms that our approach is generalizable.% Our approach is valid and effective and a more complex dataset can be effectively learned given that other, simpler data is available.

%\section{Future Works}% Do we need to have this in a 2-page summary of existing work?
%We suggest to extend the study of bias to other extents such as touristic activities and other factors that could affect the probability of recapturing different animals on social media.

Our approach needs to be tested on other social media platforms and other species. 
% TOSHORTEN It will benefit from extending the set of features available to the models. 
We expect that features accounting for the special diversity that is present in realistic datasets will improve the performance. Collecting more data will also allow for an easier detection of existing patterns in the sharing behavior, particularly in more complex and irregular datasets.
%\section{Acknowledgments}
%The research in this paper was supported in part by the NSF grants ??

\vskip 10pt

\bibliographystyle{abbrv}
\bibliography{sample-base}

\end{document}